\documentclass[letterpaper, 10 pt, conference]{IEEEtran}  
\IEEEoverridecommandlockouts   
\usepackage{graphicx} 
\usepackage{multirow}
\usepackage{amssymb}
\usepackage{stmaryrd}
\usepackage{graphicx}
\graphicspath{{Images/}}
\usepackage{epstopdf}
\usepackage{enumerate}
\usepackage{bm}
\usepackage{bbm}
\usepackage{xcolor}
\usepackage{amsmath}
\usepackage{upgreek}
\usepackage{balance}
\usepackage[ruled,linesnumbered]{algorithm2e}
\usepackage{algpseudocode}
\usepackage{cite}
\usepackage{soul}
\usepackage{nomencl}
\usepackage{hyperref}

\usepackage{makecell}
\usepackage{siunitx}
\newtheorem{remark}{Remark}

\usepackage{booktabs}
\usepackage{multirow}
\makenomenclature

\usepackage[most]{tcolorbox}
\usepackage[normalem]{ulem} 
\definecolor{promptframe}{HTML}{1F9FC4}
\definecolor{promptback}{HTML}{F2FAFD}
\definecolor{annot}{HTML}{0E7C86}
\definecolor{privacy}{HTML}{8B5FBF}
\definecolor{withheld}{HTML}{9A9A9A}

\newcommand{\pann}[1]{\textcolor{annot}{\% #1}}

\begin{document}

\title{
\LARGE \bf
Privacy-Preserving Prompted Policy Search for Robotic Control
}

\author{Ali Irshayyid, Feng Lin, Chong Li and Jun Chen, 
\IEEEmembership{Senior Member, IEEE}
\thanks{This work is supported in part by National Science Foundation through Award \#2432098 and \#2432099. \textit{J. Chen is the corresponding author. }}
\thanks{Ali Irshayyid and Jun Chen are with the Department of Electrical and Computer Engineering, Oakland University, Rochester, MI 48309 USA (e-mail: \{aliirshayyid,junchen\}@oakland.edu). Feng Lin is with the Department of Electrical and Computer Engineering, Wayne State University, Detroit, MI 48202, USA (e-mail: flin@wayne.edu). Chong Li is with OORT and the Department of Electrical Engineering, Columbia University, NYC, 10027, USA (email: cl3607@columbia.edu). }
}

\maketitle
\thispagestyle{plain} 
\pagestyle{plain}     

\begin{abstract}
Large language models (LLMs) have recently demonstrated promising capabilities as in-context policy optimizers for Reinforcement Learning (RL), enabling policy search driven by both numerical reward signals and natural language reasoning. However, deploying such methods in practice requires transmitting raw policy parameters and rewards history to cloud-based LLM APIs, exposing proprietary control strategies to third-party service providers. To address this issue, this paper introduces Privacy-Preserving Prompted Policy Search (PP-ProPS), a framework that enables LLM-guided policy optimization while keeping policy and environmental parameters confidential. PP-ProPS encodes policy parameters and reward values using secret client-side transformations before they are included in each API request, ensuring that the LLM provider observes only encoded policy parameters and scaled reward information. Furthermore, unlike Vanilla ProPS, the proposed framework does not require the true optimal episodic return to be known or disclosed to the LLM. Beyond protecting the optimization data, PP-ProPS improves the search process in two ways. First, it provides the LLM with individual reward components instead of only a single total return, offering more informative feedback about each candidate policy. Second, it uses a bounded history that prevents the prompt from growing indefinitely, improving search with high-dimensional policies and supporting the use of open-weight LLMs. The proposed PP-ProPS is evaluated on both continuous and discrete control problems spanning Multi-Joint dynamics with Contact (MuJoCo) locomotion, classic control, highway driving, and robotic arm manipulation. Compared to Vanilla ProPS, the proposed PP-ProPS outperforms ProPS in seven of the ten evaluated tasks, and surpasses conventional RL methods including PPO, SAC, and TRPO, in five of the six tasks.
\end{abstract}


\section{Introduction}

Reinforcement learning (RL) enables autonomous agents to learn control policies through interactions with an environment and has achieved notable results in strategic games \cite{silver2016mastering}, robotics \cite{wang2020mobile}, battery pack control \cite{irshayyid2026real}, and autonomous vehicles \cite{irshayyid2024review}. Traditional RL methods, however, primarily learn from numerical feedback expressed as scalar rewards \cite{sutton2018reinforcement}. In contrast, humans often combine numerical outcomes with language, prior knowledge, and common sense when learning to perform new tasks \cite{lupyan2016language}. Real-world control tasks may also be accompanied by domain descriptions, expert instructions, and operational constraints that conventional RL methods cannot directly exploit. Prior robotic-learning studies have investigated natural-language instructions for initializing RL policies and language-conditioned policy learning for long-horizon manipulation tasks \cite{tambwekar2023natural,mees2022calvin}. Such information can provide useful inductive biases by guiding exploration and providing task-specific constraints. 

Recent studies have shown that large language models (LLMs) possess capabilities that extend beyond language generation \cite{kojima2022large}. In addition to understanding natural language instructions, LLMs can recognize patterns from in-context demonstrations and reason over previously evaluated solutions \cite{mirchandani2023large}. This capability enables in-context numerical optimization, in which the problem is described through a prompt and the LLM iteratively generates candidate solutions based on previously observed solution–score pairs. Unlike gradient-based optimization, this process does not require the objective function or its derivatives to be available to the LLM. Instead, the LLM iteratively uses the performance of previously evaluated candidates to propose solutions with higher objective values.

Building on this capability, Prompted Policy Search (ProPS) \cite{zhou2025prompted} places an LLM directly within the RL policy-optimization loop. Rather than using the LLM only to design rewards \cite{turcato2025towards}, generate high-level robot task plans \cite{tsushima2025task}, or generate control actions \cite{han2024large}, ProPS uses the LLM to generate and refine the policy parameters directly. Related work has also used LLMs to generate numerical motion parameters for expressive robot behaviors \cite{roy2025gpt}. However, this approach does not iteratively optimize a control policy using feedback obtained from environment interactions. In Vanilla ProPS, at each iteration, the LLM receives a history of previously evaluated parameters and their associated episodic rewards, and then proposes a new parameter vector intended to improve policy performance. The proposed policy is executed independently in the environment, and the resulting reward is returned to the LLM as feedback for the next iteration. Vanilla ProPS additionally assumes that the true optimal episodic return is known in advance and includes this value in the prompt as a target for the optimization. This assumption may be restrictive in practical control problems, where the optimal achievable return is generally unknown and may vary with the system configuration or operating conditions. This formulation treats policy optimization as an in-context reasoning problem and allows numerical reward information to be combined with task descriptions, domain knowledge, and human-provided guidance. 

Despite these advantages, the deployment model underlying ProPS introduces a practical concern that requires attention. When a cloud-hosted LLM is used as the optimizer, the raw policy parameters and their corresponding rewards must be repeatedly included in Application Programming Interface (API) requests. In robotics, autonomous driving, and industrial control, these parameters may represent proprietary control strategies or encode information about the behavior and design of a system \cite{gummadi2024fed}. Moreover, the sequence of parameter–reward evaluations may reveal not only individual candidate policies but also how the policy evolves and which parameter configurations produce desirable behavior. Existing LLM-based policy-search methods optimize policies using scalar \cite{yang2024large, zhou2025prompted, zhang2024revisiting, amor2025sas} or trajectory-level feedback \cite{hara2026reflective} but do not consider how the optimization process can be performed when the underlying policy must remain confidential.

To address this limitation, we introduce Privacy-Preserving Prompted Policy Search (PP-ProPS), an LLM-guided policy-optimization framework that operates in an encoded parameter space. Before each LLM query, the trusted client applies a secret coordinate-wise transformation to the policy parameters and scales the reward-valued feedback using a secret transformation. The LLM therefore receives only encoded policy parameters and scaled reward information, while parameter decoding, policy execution, and environment interaction remain on the client. PP-ProPS also removes the need to provide the true optimal return and performs the search using only the feedback from previously evaluated candidates. PP-ProPS further provides component-level reward and diagnostic feedback to describe why a candidate policy succeeds or fails. A bounded-history representation is also introduced to limit prompt growth and support policy search using open-weight LLMs.
The contributions of the paper are summarized as follows. 
\begin{itemize}
    \item Unlike existing LLM-based policy-search methods that expose the original policy or control parameters \cite{zhou2025prompted,amor2025sas, hara2026reflective}, we introduce PP-ProPS, which applies secret client-side transformations to the policy parameters and reward-valued feedback. The external LLM observes only encoded parameters and scaled rewards, while the original values and encoding variables remain local.
    \item In contrast to prior LLM-based optimization methods that operate in the original parameter space \cite{yang2024large,zhou2025prompted}, we demonstrate effective policy search in an encoded parameter-space for policies containing up to 782 parameters. PP-ProPS outperforms Vanilla ProPS in seven of the ten evaluated environments while remaining comparable in the rest.
    \item Compared with the scalar episodic feedback used by ProPS \cite{zhou2025prompted}, we provide component-level reward and diagnostic feedback that improves the average policy-search performance in nine of the ten evaluated environments.
    \item Unlike existing literature methods that retain the full optimization history in each prompt \cite{zhou2025prompted,amor2025sas, hara2026reflective}, we use a bounded top-$K$ history representation that prevents prompt growth with the number of iterations. Retaining the ten highest-reward candidates reduces the final prompt size and LLM response time by 80.2\% and 74.9\%, respectively, while maintaining comparable policy-search performance and supporting policy search with a 20B-parameter open-weight LLM.
\end{itemize}

\begin{figure}[ptb]
    \centering
    \includegraphics[width=0.7\linewidth]{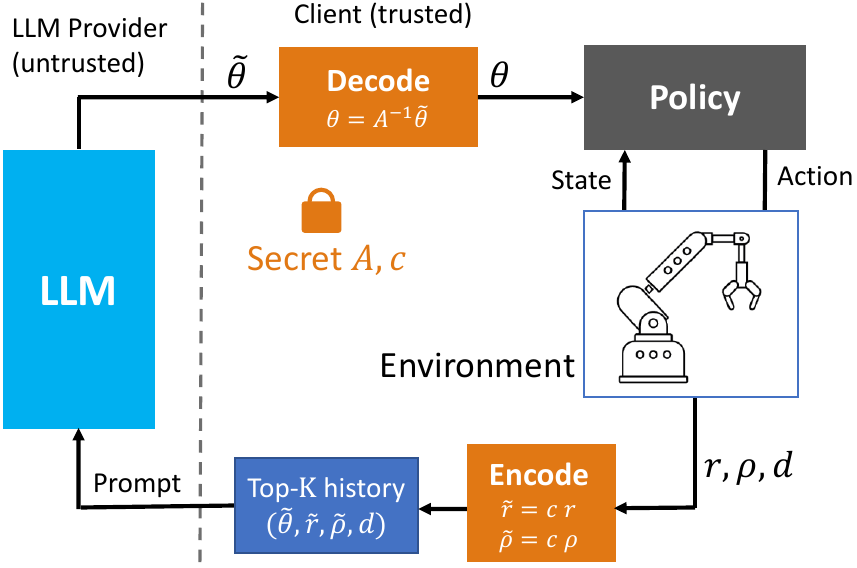}
    \caption{Overview of the proposed PP-ProPS framework. The LLM searches in the encoded parameter space $\widetilde{\theta}$, while decoding, policy execution, reward scaling, and storage of the secret variables $A$ and $c$ are performed on the trusted client.}
    \label{fig:framework}
\end{figure}

\section{Methodology}

This section first formulates Vanilla ProPS and then presents the parameter- and reward-encoding transformations, component-level reward feedback, bounded-history representation, and LLM-guided policy-search procedure.

\subsection{Problem formulation}

We consider episodic RL with a parameterized policy $\pi_\theta$, where $\theta \in \mathbb{R}^D$ denotes the policy parameters. For a trajectory $\tau$ generated by $\pi_{\theta}$ (denoted by $\tau \sim \pi_\theta$), let $R(\tau)$ denote its cumulative episodic return. The policy objective is 
\begin{equation}
    \theta^{\star} \in
    \arg\max_{\theta} J(\theta),
    \qquad
    J(\theta)=\mathbb{E}_{\tau \sim \pi_\theta}[R(\tau)].
    \label{eq:ep-return}
\end{equation}

ProPS \cite{zhou2025prompted} uses an LLM, $\mathcal{M}$, as the policy optimizer. Let $\mathcal{P}$ denote the fixed prompt information, including the task description, policy structure, parameter constraints, optimal value of $J(\theta)$, and output-format instructions. After $i$ candidate policies have been evaluated, the optimization history containing the parameter-return pair from each iteration $j=1,\ldots,i$ is defined as
\begin{equation}
    \Gamma^{(i)} =
    \left\{
    \left(\theta_{j},r_{j}\right)
    \right\}_{j=1}^{i},
    \qquad
    r_j = R(\tau_j),\quad
    \tau_j\sim\pi_{\theta_j}.
    \label{eq:props_history}
\end{equation}
At the next optimization iteration, the LLM generates a new candidate according to
\begin{equation}
    \theta_{i+1} = \mathcal{M} \left(\mathcal{P}, \Gamma^{(i)} \right),
    \label{eq:props_update}
\end{equation}
where $\mathcal{M}(\cdot)$ denotes the parameter vector extracted from the LLM response. The candidate policy is then executed in the environment, and its episodic return is appended to the history $\Gamma^{(i+1)} = \Gamma^{(i)} \cup \left\{ (\theta_{i+1},r_{i+1}) \right\}$.
Thus, ProPS performs policy search through repeated LLM queries without requiring policy gradients, a value function, or an environment model within the optimization procedure. 

To preserve privacy, the proposed PP-ProPS instead transmits only an encoded optimization history to the LLM. The original parameters $\theta$, secret encoding variables $A$ and $c$, policy execution, and all environment interactions remain on the trusted client, as illustrated in Fig. \ref{fig:framework}.


\subsection{Parameter Encoding}
\label{sec:Meth-PE}

At the beginning of each optimization run, the client generates a secret positive diagonal encoding matrix
\begin{equation}
A = \operatorname{diag}
( a_{1}, a_{2}, \ldots, a_{D} ), \quad a_j > 0,
\label{eq:encoding_matrix}
\end{equation}
where each scaling coefficient $a_j$ is independently sampled. A new matrix is generated for every independent optimization run and remains fixed across all LLM queries within that run. Maintaining the same matrix provides a consistent encoded coordinate system from which the LLM can identify relationships between parameter changes and policy performance. Before transmission, the original policy vector is encoded as
\begin{equation}
\widetilde{\theta} = A \theta,
\label{eq:parameter_encoding}
\end{equation}
where $\widetilde{\theta} \in \mathbb{R}^{D}$ denotes the encoded parameter vector observed by the LLM. Because $a_j>0$ for all $j$, the matrix $A$ is invertible. After the LLM generates an encoded candidate $\widetilde{\theta}_i$, the client recovers the corresponding policy parameters locally using
\begin{equation}
\theta_i = A^{-1}\widetilde{\theta}_i.
\label{eq:parameter_decoding}
\end{equation}

The diagonal transformation preserves the policy layout and parameter identities while introducing an unknown scaling factor for each parameter dimension. Consequently, the LLM can continue to use the task description, policy structure, and relationships observed in the optimization history, but it can never infer the true policy parameters deployed for robotic control.

\subsection{Component-Level Feedback and Reward Encoding}
\label{sec:Meth-DF}

The total episodic return provides a scalar measure of candidate policy performance but may not indicate which aspects of the behavior contribute to the observed result. Many control environments define the reward using interpretable terms associated with objectives such as forward progress, control effort, healthy behavior, collision avoidance, and task success. Providing these quantities separately allows the LLM to distinguish between policies that achieve similar total returns through different behaviors.

For each evaluated candidate, PP-ProPS therefore reports the scalar episodic return together with the available component-level reward information. Let $
\rho_i = [
\rho_{i,1},
\rho_{i,2},
\ldots,
\rho_{i,Q} ] $
denote the additional feedback associated with candidate $i$, where $Q$ is the number of reward components and $\rho_{i,q}$ represents the episodic contribution of the $q$th reward term. Task-specific diagnostic quantities vector $d_i$, such as episode length or success indicators, is reported separately. The exact feedback provided for each environment is summarized in Table \ref{tab:envs}.

Although parameter encoding prevents the direct transmission of the original policy parameters, the absolute reward magnitude may provide information regarding the operating condition or the performance of the evaluated policy relative to a known benchmark. To obscure the direct reward magnitude, the client samples a secret positive scaling factor $c$ at the beginning of each optimization run. The same factor remains fixed throughout the run and is applied to the reported reward values:
\begin{equation}
\widetilde{r}_i = c r_i, \quad c>0.
\label{eq:reward_encoding}
\end{equation}
The same scaling factor is applied to each reward-valued component:
\begin{equation}
\widetilde{\rho}_{i,q} = c\rho_{i,q},
\quad
q=1,\ldots,Q.
\label{eq:component_encoding}
\end{equation}
Because $c$ is positive, reward scaling preserves the ranking of the evaluated candidates and therefore does not change the policy selected as the best performing candidate. 
In other words, reward encoding obscures the direct absolute reward scale while retaining the information required to compare candidate policies.

\begin{remark}
Vanilla ProPS \cite{zhou2025prompted} assumes that the true optimal return is known in advance and provides this value in the prompt to guide the LLM. In the complete PP-ProPS configuration, the LLM receives no information about the true optimal return.
\end{remark}

\subsection{Bounded History Representation}
\label{sec:bounded_history}

Existing prompt-based policy-search methods append each evaluated candidate to the optimization history $\Gamma^{(i)}$ included in the prompt \cite{zhou2025prompted, amor2025sas}. Because each history entry contains a complete policy vector and its associated reward feedback, the prompt size grows with both the number of evaluated candidates and the policy dimension $D$, which may become prohibitive for long optimization runs or high-dimensional policies.

To limit this growth, the proposed PP-ProPS maintains the complete optimization encoded history locally but includes only the $K$ highest-reward candidates in each LLM prompt. Let $\widetilde{\Gamma}^{(i)}= \{(\widetilde{\theta}_j,\widetilde{r}_j, \widetilde{\rho}_j,d_j)\}_{j=1}^{i}$ denote the complete encoded history after $i$ candidate evaluations. The bounded history is obtained as
\begin{equation}
\widetilde{\Gamma}_{K}^{(i)} = \operatorname{Sort}_{\uparrow} \left( \operatorname{TopK}_{\widetilde{r}} \left( \widetilde{\Gamma}^{(i)}, \min
\left( K, \left| \widetilde{\Gamma}^{(i)} \right| \right)
\right)
\right),
\label{eq:bounded_history}
\end{equation}
where $\operatorname{TopK}(\cdot, k)$ returns the $k$ history entries with the highest encoded rewards, and $\operatorname{Sort}_{\uparrow}(\cdot)$ orders the selected entries from the lowest to the highest encoded reward such that the best-performing candidate appears last in the prompt. The minimum operator handles early iterations in which fewer than $K$ candidates have been evaluated. Because the reward scaling factor $c$ is positive, ranking candidates by $\widetilde{r}$ produces the same ordering as ranking them by the original rewards $r$. If each history entry contains $D$ encoded parameters, the history requirement of the bounded representation is $\mathcal{O}(KD)$. In comparison, an unbounded history containing $H_i$ evaluated candidates requires $\mathcal{O}(H_iD)$, where $H_i$ is the number of candidates evaluated by iteration $i$. Therefore, the bounded representation maintains a fixed history size in the prompt as the number of optimization iterations increases.

\subsection{Privacy-Preserving Prompted Policy Search}
\label{sec:pp-props}
Algorithm \ref{alg:pp_props} summarizes the complete PP-ProPS combining prompt policy search, parameters and rewards encoding, component-level reward feedback, and bounded history representation. At the beginning of each optimization run, the client generates the secret parameter-encoding matrix $A$ and reward scaling factor $c$, which remain fixed throughout the run. Let $N_w$ and $N_o$ denote the numbers of randomly sampled warm-up policies and LLM-guided optimization iterations, respectively, and let $\mathcal{E}$ denote the execution environment. The first $N_w$ policies are sampled from the feasible parameter range and evaluated to initialize the encoded history $\widetilde{\Gamma}$. Each history entry contains the encoded policy parameters, scaled scalar and component-level rewards, and any non-reward diagnostic feedback. The original returns remain local and are used only to identify the best performing policy. 

At iteration $i$, the client retains $K$ candidates with the highest encoded returns and orders them from lowest to highest return. This history is combined with the task description, policy structure, and behavioral guidance to form the prompt $\mathcal{P}_i$ (illustrated in Fig. \ref{fig:PP-Prpos-prompt-structure}). The LLM generates a candidate $\widetilde{\theta}_i$ in the encoded space, which the client decodes using \eqref{eq:parameter_decoding}, evaluates in the environment, and appends the resulting encoded parameter, reward, component-feedback, and diagnostic tuple to $\widetilde{\Gamma}$. Unlike ProPS \cite{zhou2025prompted}, the prompt does not disclose the true optimal achievable return. Note that the privacy-preserving transformation is only needed during optimization. Once the (decoded) best policy $\pi_{\theta_{\textit{best}}}$ is found, it can be deployed for robot control directly without requiring any encoding during execution.

\begin{figure}[ptb]
\centering
\begin{tcolorbox}[
    colback=promptback,
    colframe=promptframe,
    coltitle=white,
    fonttitle=\bfseries,
    title={PP-ProPS Prompt},
    boxrule=0.8pt,
    arc=2pt,
    left=5pt,
    right=5pt,
    top=4pt,
    bottom=0pt
]
\footnotesize
You are a good global RL policy optimizer, helping me find the global optimal policy in the following environment:

\smallskip
\setlength{\tabcolsep}{0pt}
\begin{tabular}{@{}p{0.42\linewidth}@{\hspace{0.5em}}p{0.54\linewidth}@{}}
1. The task description:
    & \pann{Environment, state and action semantics, policy structure, and hints} \\

2. The policy parameters:
    & \pann{parameters are encoded} \\[2pt]
3. The feedback: & \pann{scaled return and component-level terms} \\
4. How we will interact:
    & \pann{formatting and output instructions} \\[2pt]

5. Here is the history:
    & \pann{bounded top-$K$, best entry last}

\end{tabular}
\end{tcolorbox}

\caption{Summary of the PP-ProPS prompt structure.}
\label{fig:PP-Prpos-prompt-structure}
\end{figure}

\begin{algorithm}[ptb]
\caption{Privacy-Preserving Prompted Policy Search (PP-ProPS)}
\label{alg:pp_props}

Initialize
$A$, $a_j>0$, $c>0$, $\widetilde{\Gamma} \gets \emptyset$,
$r_{\textit{best}}\gets-\infty$, and
$\theta_{\textit{best}}\gets\emptyset$\;

\For{$m=1$ to $N_w$}{
    Sample $\theta_m$ from the feasible parameter range\;
    Execute $\pi_{{\theta}_m}$ in $\mathcal{E}$ and obtain
    $r_m$, $\rho_m$, and $d_m$\;
    $\widetilde{\theta}_m\gets A \theta_m$\;
    $\widetilde{r}_m\gets cr_m$ and
    $\widetilde{\rho}_m\gets c \rho_m$\;
    $\widetilde{\Gamma} \gets\widetilde{\Gamma}\cup
    \left\{
    (\widetilde{\theta}_m,\widetilde{r}_m,
    \widetilde{\rho}_m,d_m)
    \right\}$\;
    \If{$r_m>r_{\textit{best}}$}{
        $r_{\textit{best}}\gets r_m$ and
        $\theta_{\textit{best}}\gets \theta_m$\;
    }
}

\For{$i=1$ to $N_o$}{
    $K_i\gets\min(K,|\widetilde{\Gamma})$\;
    $\widetilde{\Gamma}_{K}
    \gets
    \operatorname{Sort}^{\uparrow}
    \left(
    \operatorname{TopK}_{\widetilde{r}}
    (\widetilde{\Gamma},K_i)
    \right)$\;
    Construct $\mathcal{P}_i$ using $\widetilde{\Gamma}_{K}$\;
    $\widetilde{\theta}_i\gets\mathcal{M}(\mathcal{P}_i)$\;
    $\theta_i\gets A^{-1}\widetilde{\theta}_i$\;
    Execute $\pi_{\theta_i}$ in $\mathcal{E}$ and obtain
    $r_i$, $\rho_i$, and $d_i$\;
    $\widetilde{r}_i\gets cr_i$ and
    $\widetilde{\rho}_i\gets c \rho_i$\;
    $\widetilde{\Gamma} \gets \widetilde{\Gamma} \cup
    \left\{
    (\widetilde{\theta}_i,\widetilde{r}_i,
    \widetilde{\rho}_i,d_i)
    \right\}$\;
    \If{$r_i>r_{\textit{best}}$}{
        $r_{\textit{best}}\gets r_i$ and
        $\theta_{\textit{best}}\gets \theta_i$\;
    }
}

Return $\theta_{\textit{best}}$\;
\end{algorithm}

\section{Experimental Setup}

The experiments evaluate four aspects of PP-ProPS: the effect of parameter encoding, the contribution of component-level feedback, the performance of the complete parameter- and reward-encoded formulation, and the tradeoff between bounded-history performance and prompt size. The LLM-guided configurations are also compared with standard RL algorithms.

\begin{figure}[ptb]
    \centering
    \includegraphics[width=0.7\linewidth]{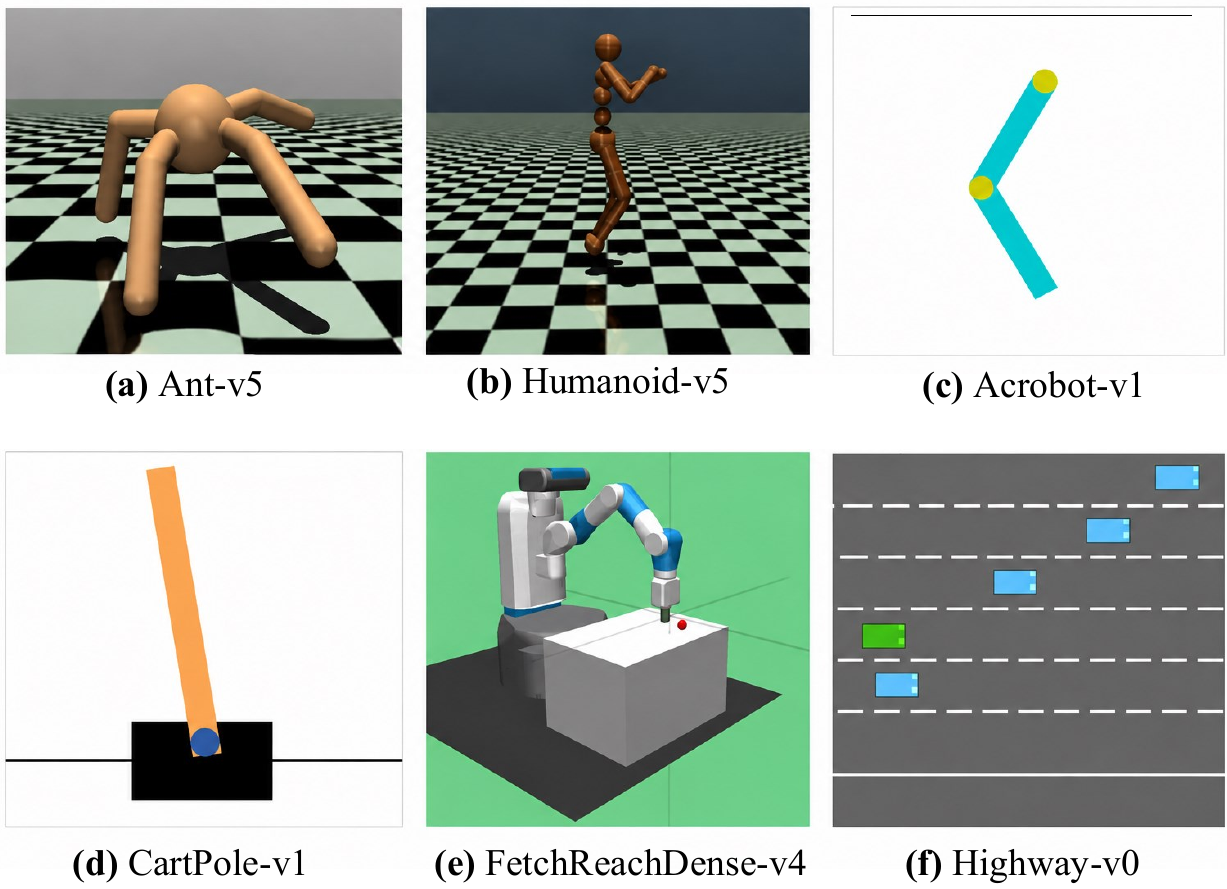}
    \caption{Environments used to evaluate the proposed PP-ProPS framework.}
    \label{fig:Envs}
\end{figure}

We consider ten environment configurations spanning four domains, as shown in Fig. \ref{fig:Envs} and summarized in Table \ref{tab:envs}, MuJoCo locomotion \cite{todorov2012mujoco} (Ant-v5 \cite{ant}, Humanoid-v5 \cite{humanoid}), classic control tasks (Acrobot-v1 \cite{acrobot}, CartPole-v1 \cite{cartpole}), robotic arm manipulation (FetchReachDense-v4 \cite{arm}), and highway driving (highway-v0) \cite{Leurent_An_Environment_for_2018}. The environments include both continuous and discrete action spaces. For Ant-v5 and Humanoid-v5, three observation configurations, denoted as Ultra, Joints, and Torso, progressively expand the observation vector while preserving the underlying control task. Ultra contains selected body, joint, and velocity variables; Joints adds locomotion-related joint states; and Torso contains the complete generalized position and velocity states. The corresponding policy dimensions range from 88 for Ant-Ultra to 782 for Humanoid-Torso.

A linear policy is adopted for all evaluated environments, consistent with benchmark Vanilla ProPS \cite{zhou2025prompted}. For an environment with $n_s$ observations and $n_a$ actions, the policy is defined as
\begin{equation}
\mathbf{u} = W^{\top}\mathbf{s} + \mathbf{b},
\label{eq:linear_policy}
\end{equation}
where $\mathbf{s}\in\mathbb{R}^{n_s}$ denotes the observation vector, $W\in\mathbb{R}^{n_s\times n_a}$ is the policy weight matrix, $\mathbf{b}\in\mathbb{R}^{n_a}$ is the bias vector, and $\mathbf{u}\in\mathbb{R}^{n_a}$ is the policy output. For continuous-action environments, $\mathbf{u}$ is applied as the actuation vector subject to the environment action limits. For discrete-action environments, the elements of $\mathbf{u}$ are treated as action scores, and the action with the highest score is executed. The resulting policy-parameter vector has dimension $D = (n_s+1)n_a$.
To isolate the contribution of each mechanism introduced, four configurations are evaluated in this study as follows.
\begin{itemize}
    \item \textbf{Vanilla ProPS} \cite{zhou2025prompted}: In this configuration, the LLM observes the true parameters $\theta$, and each candidate policy performance is evaluated by its scalar episodic reward. The prompt discloses the reward function structure and the true maximum achievable return.
    \item \textbf{Parameter-Encoded ProPS (PE-ProPS)}: This configuration isolates the effect of parameter encoding. The task description, policy structure, and scalar reward feedback are identical to those of Vanilla ProPS. However, each policy vector is encoded using the secret diagonal transformation $A$ in \eqref{eq:parameter_encoding} before being included in the prompt. The original policy parameters remain local to the client.
    \item \textbf{Detailed-Feedback ProPS (DF-ProPS)}: This configuration evaluates the contribution of additional policy-performance information without applying parameter or reward encoding. The LLM observes the original policy parameters and scalar episodic return, together with the component-level rewards and task-specific feedback summarized in Table \ref{tab:envs}.
    \item \textbf{Privacy-Preserving Prompted Policy Search (PP-ProPS)}: This configuration represents the complete proposed framework. The policy parameters are encoded according to \eqref{eq:parameter_encoding}, while the scalar episodic return and the reward-valued feedback terms are scaled using the secret factor defined in \eqref{eq:reward_encoding}. The detailed feedback used by DF-ProPS is retained, whereas the maximum achievable return is withheld from the prompt.
\end{itemize}
Reflective Prompted Policy Optimization (R2PO) \cite{hara2026reflective} is also evaluated as a recent baseline in which a Search-LLM proposes initial policy, a Critic-LLM revises it using trajectory evidence and the higher-return candidate is retained. R2PO uses the same 10-policy random warm-up and 25 two-stage iterations to match the 50 LLM calls used by the ProPS-based configurations. Following \cite{hara2026reflective}, each initial and revised
candidate is evaluated over 20 episodes.

All LLM-guided policy-search configurations use \texttt{gpt-oss-20b} open-weight reasoning model \cite{agarwal2025gpt}, with identical inference settings across all configurations. For PE-ProPS and PP-ProPS, a new encoding matrix $A$ is independently generated at the beginning of each run and remains fixed throughout that run. PP-ProPS additionally generates a new reward-scaling factor $c$ at the beginning of each run. Unless otherwise stated, all ProPS-based configurations use the bounded top-$K$ history with $K=10$. The bounded and unbounded history-formulations are compared in Section \ref{sec:res-b-hist}.

In addition, seven standard RL algorithms are evaluated using Stable-Baselines3 \cite{stable-baselines3} default hyperparameters: Proximal Policy Optimization (PPO) \cite{schulman2017proximal}, Advantage Actor-Critic (A2C) \cite{mnih2016asynchronous}, Trust-Region Policy Optimization (TRPO) \cite{schulman2015trust}, Soft Actor-Critic (SAC) \cite{haarnoja2018soft}, Twin-Delayed Deep Deterministic Policy Gradient (TD3) \cite{fujimoto2018addressing}, Deep Deterministic Policy Gradient (DDPG) \cite{lillicrap2015continuous}, and Deep Q-Network (DQN) \cite{mnih2015human}. SAC, TD3, and DDPG are applied only to environments with continuous action spaces, whereas DQN is applied only to environments with discrete action spaces. Inapplicable combinations are indicated by ``--'' in Table \ref{tab:baselines}. The Ant-v5 and Humanoid-v5 baseline results reported in Table \ref{tab:baselines} use the Ultra observation configuration.

\begin{table*}[ptb]
\centering
\caption{Evaluated environments, policy dimensions, and detailed feedback provided to the LLM}
\label{tab:envs}
\begin{tabular}{lcccp{7.1cm}c}
\toprule
Environment
& Action type
& $n_a$
& $n_s$
& Detailed feedback
& $D$ \\
\midrule

Ant-v5 (Ultra)
& Continuous
& 8
& 10
& Forward-progress reward, healthy reward, control-cost penalty, and contact-force penalty
& 88 \\ 

Ant-v5 (Joints)
& Continuous
& 8
& 16
& Same as Ant-v5 (Ultra)
& 136 \\

Ant-v5 (Torso)
& Continuous
& 8
& 27
& Same as Ant-v5 (Ultra)
& 224 \\

Humanoid-v5 (Ultra)
& Continuous
& 17
& 11
& Forward-progress reward, healthy reward, control-cost penalty, and contact-force penalty
& 204 \\

Humanoid-v5 (Joints)
& Continuous
& 17
& 26
& Same as Humanoid-v5 (Ultra)
& 459 \\

Humanoid-v5 (Torso)
& Continuous
& 17
& 45
& Same as Humanoid-v5 (Ultra)
& 782 \\

Acrobot-v1
& Discrete
& 3
& 6
& Episode length, task-success indicator, and average height of the free-tip
& 21 \\

CartPole-v1
& Discrete
& 2
& 4
& Episode length, and termination cause
& 10 \\

FetchReachDense-v4
& Continuous
& 4
& 6
& Episode length and task-success indicator
& 28 \\

Highway-v0
& Discrete
& 5
& 15
& High-speed reward, right-lane reward, collision penalty, and on-road reward
& 80 \\

\bottomrule
\end{tabular}
\end{table*}

\begin{figure}[ptb]
    \centering
    \includegraphics[width=0.8\linewidth]{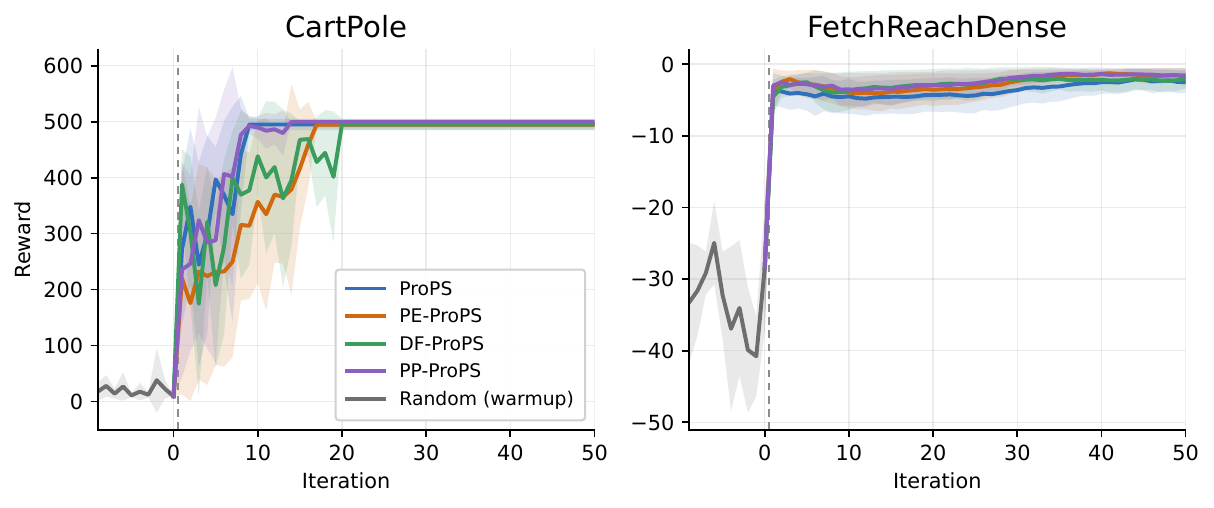}
    \caption{Optimization performance of the evaluated ProPS configurations across two environments. Curves represent the mean episodic rewards and shaded regions indicate the standard deviation over five independent runs. Iterations preceding $i=0$ correspond to the 10-policy random warm-up phase.}
    \label{fig:optimization_performance}
\end{figure}
\section{Results}

This section evaluates the bounded-history representation, the effect of parameter encoding, the contribution of component-level feedback, and performance relative to R2PO and standard RL algorithms. Unless otherwise stated, results are reported in Table \ref{tab:all_results} as the mean episodic return of the best policy found under each method's stated evaluation protocol, with the standard deviation across five independent runs shown in parentheses. All LLM-based methods use a 10-policy random warm-up. The best result for each environment is highlighted in bold. 

Fig. \ref{fig:optimization_performance} illustrates the reward curves for two representative environments. Following the random warm-up phase, all four configurations rapidly approach the maximum CartPole return of 500, whereas improvement is more gradual for FetchReachDense. PP-ProPS follows trends comparable to the unencoded configurations despite receiving only encoded policy parameters and rewards. 

\begin{table*}[ptb]
\centering
\caption{Comparison of Episodic Rewards for Vanilla ProPS, R2PO, PE-ProPS, DF-ProPS, and PP-ProPS across all evaluated environments}
\label{tab:all_results}
\begin{tabular}{lccccc}
\toprule
Environment & Vanilla ProPS & R2PO & PE-ProPS & DF-ProPS & PP-ProPS\\
\midrule

Ant-Ultra
& 942.75 (59.11)
& Context overflow
& 998.64 (4.75)
& \textbf{998.9} (37.5)
& 994.55 (10.72) \\

Ant-Joints
& 966.58 (48.61)
& Context overflow
& 982.62 (26.45)
& \textbf{996.15} (2.79)
& 989.68 (17.11) \\

Ant-Torso
& 929.19 (61.73)
& Context overflow
& 982.41 (28.62)
& 993.76 (0.27)
& \textbf{997.08} (7.69) \\

Humanoid-Ultra
& 448.2 (51.65)
& 443.7 (80.5)
& 382.51 (73.79)
& \textbf{475.91} (72.61)
& 375.52 (96.64) \\

Humanoid-Joints
& \textbf{394.38} (90.88)
& 293.2 (77.1)
& 364.5 (94.03)
& 387 (69.3)
& 361.59 (97.2) \\

Humanoid-Torso
& 307.38 (63.7)
& 260.92 (20.74) $^\dagger$
& 251 (54.93)
& \textbf{326.43} (87.67)
& 290.68 (43.57) \\

Acrobot
& -77.6 (1.39)
& -83.5 (4.6)
& -77.4 (0.2)
& \textbf{-71.6} (0.4)
& -73.3 (1.12) \\

CartPole
& 495 (10)
& \textbf{500} (0.0)
& 494.8 (6.5)
& 495.2 (10)
& 499.9 (0.2) \\

FetchReachDense
& -0.6 (0.03)
& -0.61 (0.07)
& -0.78 (0.06)
& \textbf{-0.51} (0.05)
& -0.54 (0.03) \\

Highway Env
& 122.57 (24.25)
& 146.2 (1.4)
& 90.04 (18.86)
& 151.46 (4.16)
& \textbf{154.55} (10.33) \\

\bottomrule
\multicolumn{6}{l}{\footnotesize
$^{\dagger}$Three runs terminated early because of context overflow. The best pre-overflow policies are reported.}
\end{tabular}
\end{table*}

\begin{figure}[ptb]
    \centering
    \includegraphics[width=0.6\linewidth]{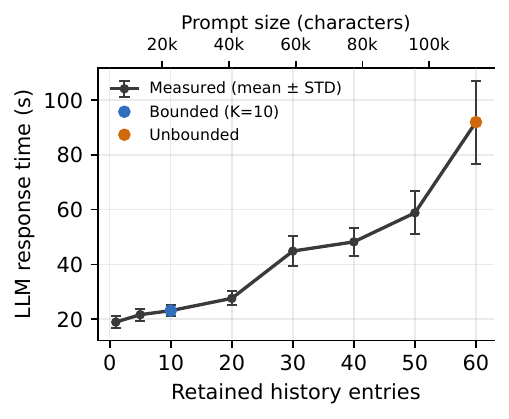}
    \caption{LLM response time for Ant-Ultra ($D = 88$) versus retained history size and the corresponding prompt size. Highlighted points denote bounded ($K=10$) and unbounded history representations.}
    \label{fig:history_cost}
\end{figure}

\subsection{Bounded-History Analysis} \label{sec:res-b-hist}

Using a 20B-parameter open-weight LLM makes prompt size a practical concern, particularly for policies containing up to 782 parameters, more than seven times the dimension evaluated in Vanilla ProPS \cite{zhou2025prompted}. Table \ref{tab:bounded_history_performance} compares the bounded and unbounded Vanilla ProPS history approaches for the three Ant policies. The unbounded history improves the return by only 0.84\%, 0.03\%, and 0.38\% for $D=88$, 136, and 224, respectively. Thus, the top-$K$ representation maintains performance comparable to the complete history.

Such a slight performance degradation, however, is rewarded by significant computation gain. As shown in Fig. \ref{fig:history_cost}, the top-10 history reduces the final prompt size from 114,000 to 22,000 characters and the mean LLM response time by 74.9\%, corresponding to an 80.2\% prompt size reduction. Accordingly, the bounded top-$K$ representation with $K=10$ is used for all other LLM-guided evaluations, unless otherwise stated.

\begin{table}[t]
\centering
\caption{Comparison of the bounded and unbounded history representations across the Ant policy dimensions}
\label{tab:bounded_history_performance}
\begin{tabular}{lccc}
\toprule
$D$
& Bounded history ($K=10$)
& Unbounded history & Diff (\%) \\
\midrule


88
& 942.75 (59.11)
& \textbf{950.75} (32.1) & $-0.84$ \\ 

136
& 966.58 (48.61)
& \textbf{966.9} (15.7) & $-0.03$\\ 

224
& 929.19 (61.73)
& \textbf{932.7} (53.7) & $-0.38$\\
\bottomrule
\end{tabular}
\end{table}

\subsection{Effect of Parameter Encoding}
\label{sec:res-encoding}

The Vanilla ProPS and PE-ProPS columns of Table \ref{tab:all_results} isolate the effect of parameter encoding. Encoding improves all three Ant configurations, increasing their average returns by approximately 1.7-5.9\% while generally reducing run-to-run variability. For Ant-Ultra, the return increases from 942.75 to 998.64 while the standard deviation decreases from 59.11 to 4.75. These results show that the LLM can identify useful parameter-reward relationships without observing the original parameter magnitudes. We hypothesize that encoding reduces the influence of pretrained assumptions about parameter magnitudes. Under Vanilla ProPS, the LLM may associate particular policy-parameter values with expected control behavior based on its prior knowledge of the environment or related dynamical systems. Such prior associations may bias candidate generation when the relationship between parameter magnitude and episodic return differs from those expectations. The encoded representation encourages the LLM to rely primarily on the numerical relationships observed during the optimization process.

For Humanoid, encoding reduces the average return across all three observation configurations. Nevertheless, PE-ProPS continues to generate effective policies for dimensions up to 782, indicating that encoded in-context policy search remains feasible at the evaluated scale.
CartPole and Acrobot are nearly unchanged, whereas the average returns for FetchReachDense and Highway environments decrease from $-0.60$ to $-0.78$ and 122.57 to 90.04, respectively.  These results suggest that, after parameter magnitudes are obscured, scalar feedback may be insufficient for tasks requiring more informative behavioral guidance, motivating the component-level feedback evaluated next.


\subsection{Effect of Component-Level Reward Feedback} \label{sec:res-subreward}


The Vanilla ProPS and DF-ProPS columns of Table \ref{tab:all_results} show that component-level and diagnostic feedback improves the average return in nine of the ten environments. For the three Ant configurations, the improvements range from 3.1\% to 6.9\%, with the standard deviations decreasing to 2.79 and 0.27 for Ant-Joints and Ant-Torso, respectively. DF-ProPS also improves Humanoid-Ultra and Humanoid-Torso by approximately 6.2\%, while the small decrease for Humanoid-Joints remains limited relative to the reported variation. The feedback, summarized in Table \ref{tab:envs}, provides information regarding forward motion, control effort, health, and contact penalties, allowing the LLM to distinguish between policies that obtain similar total returns through different behaviors.

The benefit of detailed feedback is also evident in the remaining control environments. For Acrobot, the average return improves from $-77.60$ to $-71.60$. For CartPole, the return increases from 495 to 495.2. The FetchReachDense return improves from $-0.60$ to $-0.51$. A substantial improvement is observed for Highway, where component-level feedback increases the average return from 122.57 to 151.46, corresponding to an improvement of approximately 23.6\%. 

\subsection{Performance of Complete PP-ProPS}
The final column of Table \ref{tab:all_results} evaluates the complete PP-ProPS configuration, in which the policy parameters and reward-valued feedback are encoded, component-level feedback is provided, and the optimal achievable return is withheld from the prompt (unlike Vanilla ProPS). Despite these additional privacy mechanisms, PP-ProPS outperforms Vanilla ProPS in seven of the ten evaluated environments. The largest gains occur for Ant-Torso and Highway, which improve by approximately 7.3\% and 26.1\%, respectively. PP-ProPS also improves Acrobot and FetchReachDense and achieves a CartPole return of 499.9 with the lowest variability among the four configurations. 

Performance decreases for the three Humanoid configurations, suggesting that the effectiveness of the complete encoded formulation also depends on the task dynamics and informativeness of the available feedback. Among the seven tasks with reported R2PO results, PP-ProPS outperforms R2PO in five. For Ant, the long episode horizon causes the trajectory prompt to exceed the context window before the first Critic-LLM query. Additionally, three of the five Humanoid-Torso runs also terminate before completing all iterations, and their reported returns use the best policy found before the overflow. In contrast, PP-ProPS completes all evaluations without exposing the original parameters or reward scale.

\subsection{Standard RL Baselines} \label{sec:res-rl}
\begin{table*}[t]
\centering
\caption{Performance Comparison of the Proposed PP-ProPS and Standard RL Baselines}
\label{tab:baselines}
\renewcommand{\arraystretch}{1.08}
\setlength{\tabcolsep}{5pt}
\begin{tabular}{lcccccc}
\hline
Approach & Ant & Humanoid & Acrobot & CartPole & Highway & FetchReachDense \\
\hline
PPO & \textbf{1000.16} (15.06) & 192.94 (16.38) & -303.95 (170.33) & 127.16 (179.77) & 107.01 (26.51) & -7.65 (0.46) \\
A2C & 233.61 (126.42) & 118.82 (75.77) & -209.57 (158.84) & 161.69 (177.61) & 73.68(33.65) & -12.80 (8.19) \\
SAC & -52.42 (46.51) & 78.24 (3.10) & -- & -- & -- & -8.55 (0.99) \\
TD3 & 938.04 (240.49) & 81.86 (34.78) & -- & -- & -- & -8.06 (1.79) \\
DDPG & -6.55 (72.99) & 89.59 (40.25) & -- & -- & -- & -7.39 (2.45) \\
TRPO & 965.23 (10.99) & 193.21 (16.80) & -284.65 (171.60) & 127.04 (187.54) & 109.09 (30.69) & -8.76 (0.93) \\
DQN & -- & -- & -408.80 (150.13) & 12.51 (5.98) & 39.14 (4.31) & -- \\
PE-ProPS & 998.64 (4.75) & 382.51 (73.79) & -77.4 (0.2) & 494.8 (6.5) & 90.04 (18.86) & -0.78 (0.06) \\
DF-ProPS & 998.90 (37.50) & \textbf{475.91} (72.61) & \textbf{-71.6} (0.4) & 495.2 (10) & 151.46 (4.16) & \textbf{-0.51} (0.05) \\
PP-ProPS & 994.55 (10.72) & 375.52 (96.64) & -73.3 (1.12) & \textbf{499.9} (0.2) & \textbf{154.55} (10.33) & -0.54 (0.03) \\
\hline
\end{tabular}
\end{table*}

Table \ref{tab:baselines} compares the proposed PP-ProPS with seven standard RL algorithms. For Ant-Ultra, PPO achieves the highest return of 1000.16, while DF-ProPS, PE-ProPS, and PP-ProPS remain within approximately 0.6\% of this result. Furthermore, PE-ProPS achieves a standard deviation of 4.75, compared with 15.06 for PPO. Therefore, the proposed PP-ProPS configurations maintain comparable performance while directly optimizing a compact linear policy.

The proposed PP-ProPS approaches outperform all evaluated RL baselines for Humanoid-Ultra, Acrobot, CartPole, Highway, and FetchReachDense. DF-ProPS more than doubles the best RL-baseline return for Humanoid-Ultra, while PP-ProPS improves the Highway return by approximately 41.7\% relative to TRPO. For FetchReachDense, DF-ProPS achieves $-0.51$, compared with $-7.39$ for DDPG, the strongest evaluated RL baseline. Thus, the LLM-guided configurations remain competitive on Ant-Ultra and provide substantially higher returns across all remaining continuous- and discrete-control tasks. Furthermore, PP-ProPS preserves strong policy-search performance while maintaining the privacy-preserving parameter and reward representations.

\section{Conclusion}

This paper presents Privacy-Preserving Prompted Policy Search (PP-ProPS), an LLM-guided policy-optimization framework that avoids transmitting raw policy parameters and reward information to an external LLM provider. The proposed framework combines client-side parameter and reward transformations with component-level feedback and a bounded-history representation. Evaluation across ten continuous- and discrete-control configurations demonstrates that PP-ProPS outperforms Vanilla ProPS in seven environments and remains effective for policies containing up to 782 parameters, while reducing the final prompt size and mean LLM response time by approximately 80.2\% and 74.9\%, respectively. Furthermore, the proposed framework outperforms the evaluated reinforcement learning baselines in five of the six task groups and degrades by only 0.6\% for the remaining one. 
Future work will evaluate nonlinear control policies with physical experiments and investigate nonlinear encoding against parameter-recovery and information-inference attacks.

\balance
\bibliographystyle{IEEEtran}
\bibliography{ref}
\end{document}